\documentclass[10pt,letterpaper]{article}
\usepackage{matrixreward-preprint,times}

\usepackage{amsmath,amsfonts,bm}

\def\eqref#1{equation~\ref{#1}}

\def\1{\bm{1}}

\DeclareMathAlphabet{\mathsfit}{\encodingdefault}{\sfdefault}{m}{sl}
\SetMathAlphabet{\mathsfit}{bold}{\encodingdefault}{\sfdefault}{bx}{n}

\usepackage{float}
\usepackage{graphicx}
\usepackage{algorithm}
\usepackage{algpseudocode}
\usepackage{needspace}
\usepackage{booktabs}
\usepackage{url}
\usepackage[hidelinks]{hyperref}
\hypersetup{
  pdftitle={MatrixReward: Reward from Rubric Matrix for Open-Ended Generation},
  pdfauthor={Zihan Shen, Qi Liu, Zixuan Yang, Yiqun Chen, Chenglong Zhao, Xiaozhao Wang, Lei He}
}
\newcommand{\method}{MatrixReward}
\title{\method{}: Reward from Rubric Matrix for Open-Ended Generation}
\author{%
\begin{minipage}{\textwidth}
\fontsize{10}{12}\selectfont\raggedright
\mbox{\textbf{Zihan Shen}$^{1,3,*}$},\allowbreak\hspace{0.5em}
\mbox{\textbf{Qi Liu}$^{2,*}$},\allowbreak\hspace{0.5em}
\mbox{\textbf{Zixuan Yang}$^{2}$},\allowbreak\hspace{0.5em}
\mbox{\textbf{Yiqun Chen}$^{2}$},\allowbreak\hspace{0.5em}
\mbox{\textbf{Chenglong Zhao}$^{3}$},\allowbreak\hspace{0.5em}
\mbox{\textbf{Xiaozhao Wang}$^{3}$},\allowbreak\hspace{0.5em}
\mbox{\textbf{Lei He}$^{3,\dagger}$}\\[6pt]
\mbox{$^{1}$Zhejiang University}\quad
\mbox{$^{2}$Renmin University of China}\quad
\mbox{$^{3}$Qwen Business Unit of Alibaba}
\end{minipage}}
\date{}
\begin{document}

\maketitle
\begingroup
\renewcommand{\thefootnote}{\fnsymbol{footnote}}
\footnotetext[1]{Equal contribution.}
\footnotetext[2]{Corresponding author.}
\endgroup

\begin{abstract}
Open-ended query generation lacks standard answers, thus necessitating an effective reward mechanism. Pointwise scoring rubrics provide limited information about the relative quality of sample answers under the same prompt; merging multiple rubric judgments into a single score may also mask the differences between these answers. We propose \textbf{MatrixReward}, which constructs rewards from a rollout-by-rubric win-rate matrix obtained by comparing every pair of sampled responses under each rubric. The spread of each matrix column captures how strongly that rubric distinguishes the current rollouts, while correlations between columns reveal rubric repetition; together, these statistics yield data-dependent rubric weights. We combine these weights with the prior weights of rubrics. After column normalization and weighting, the observed per-rubric maxima and minima define positive and negative ideal profiles. Each rollout’s distances to these two ideals determine its relative-closeness quality reward.  Evaluated using Qwen3-8B on four open-ended query-answering benchmarks, MatrixReward achieves an average score of 63.02, outperforming the strongest baseline by approximately 2.0\%. These results support the idea that matrices derived from relative comparisons can be used to construct rewards more reasonably for open-ended generative reinforcement learning.\footnote{Code and materials required to reproduce our results are available at \url{https://github.com/Backsingles/MatrixReward}.}
\end{abstract}

\section{Introduction}
\label{sec:introduction}

Reinforcement learning has become an important approach to improving language
models beyond their initial capabilities, but its effectiveness depends on the
quality of the reward signal
~\citep{ouyang2022training,dubois2023alpacafarm,shao2024deepseekmath}. In
verifiable tasks, such as mathematics and programming, a correct answer or
successful test provides a concrete target. In open-ended generation, feedback
must instead account for instruction following, factual grounding, relevance,
and presentation ~\citep{zheng2023judging,gunjal2026rubrics}. A response may be
fluent yet uninformative, or detailed yet poorly grounded. Effective rewards
must distinguish these qualities among responses to the same prompt.

Rubric-based rewards offer a promising foundation by expressing these
requirements as explicit evaluation rubrics
~\citep{gunjal2026rubrics,zhang2026chasing,liu2026openrubrics}. Rather than
leaving quality entirely to a holistic judgment, rubrics identify the aspects
that an evaluator should consider. However, clear
rubrics do not automatically produce informative training signals: an
evaluator must still determine how well each response satisfies them
~\citep{zheng2023judging,tripathi2025pairwise}, and those assessments must
ultimately guide a policy update. The distinction matters for group-relative
policy optimization (GRPO) ~\citep{shao2024deepseekmath}: the training signal depends on differences among sampled responses, not merely on
whether each response appears satisfactory in isolation.

Many rubric-based reward systems assign pointwise scores to responses
independently. Although rubrics specify what to judge, a numerical score lacks
a stable anchor when a query has no reference answer; the same value may
reflect different quality standards across queries or evaluators
~\citep{zheng2023judging,dubois2024length}. Rubric weights is also
commonly set when the rubrics are created. These weights represent the factors that should be considered in the response, but they cannot guarantee the differentiation between rollouts sampled during training; and if the rubric is highly correlated, it may lead to duplicate scoring. More importantly, a linearly weighted average of rubrics may cause scores across different dimensions to cancel each other out, assigning the same reward to responses of different quality under the same rubric.

\begin{figure}[t]
    \centering
    \includegraphics[width=\linewidth]{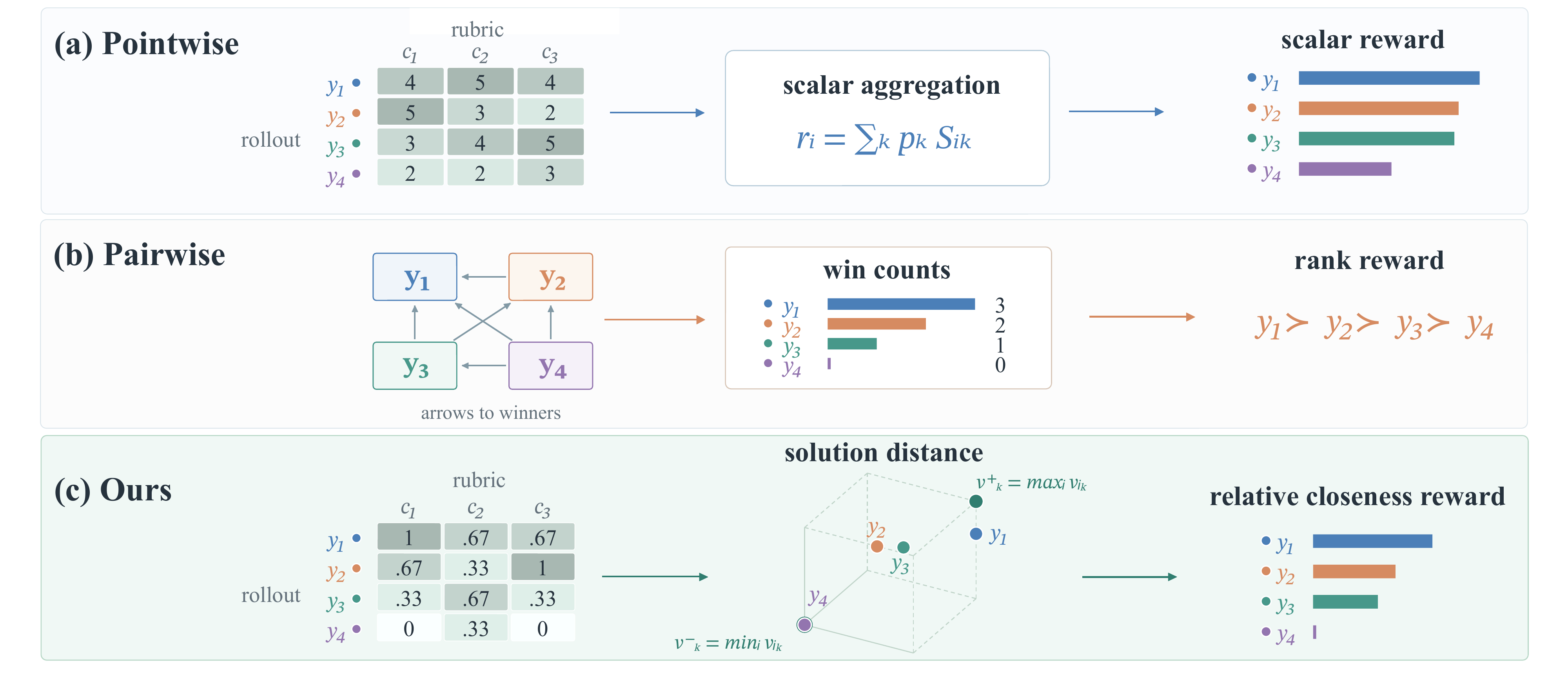}
    \caption{Comparison of reward construction for the same rollout group.}
    \label{fig:intro-comparison}
\end{figure}

To address these problems, we propose \method{}. As illustrated in
Figure~\ref{fig:intro-comparison}, rubric-wise pairwise judgments retain a
separate win-rate profile for each response. The matrix reveals which rubrics
distinguish the current rollouts and which provide overlapping information.
We combine this evidence with query-specific weights, then compare
the resulting profiles with positive and negative ideals to construct rewards.

Our contributions are as follows:
\begin{itemize}
    \item \textbf{Rubric-wise comparison matrix.} We introduce \method{}, a
    reward framework for open-ended group-based RL. By comparing every pair of
    rollouts under each query-specific rubric, it constructs a
    rollout-by-rubric win-rate matrix that retains each response's relative
    quality.
    \item \textbf{Distance-based reward construction.} We derive group-specific
 rubric weights from column dispersion and cross-rubric correlation, then
    combine them with query-specific importance priors. After normalizing and weighting the matrix columns, we use their
    observed maxima and minima as positive and negative ideal profiles. For
    each rollout, the Euclidean distance to the negative ideal divided by the
    sum of its distances to both ideals gives a scalar quality reward, favoring profiles near the positive ideal and far from the negative ideal.
    \item \textbf{Four-benchmark evaluation.} Training Qwen3-8B with \method{}
    yields a 63.02 average across four benchmarks, exceeding the strongest
    pairwise, pointwise, and weight-aware baselines by 1.23, 6.93, and 5.60
    points, respectively. Ablations examine weight extraction and fusion,
    ideal-profile selection, distance metrics, and judging cost.
\end{itemize}

\section{Related Work}

\paragraph{Rubric-based reward evaluation.}
Rubrics make the standards for judging open-ended responses explicit, but do not
by themselves establish a common numeric scale across queries or evaluators.
Rubrics as Rewards (RaR) uses instance-specific rubrics in on-policy RL, with
both weighted rubric-level assessments and rubric-conditioned holistic
judgments ~\citep{gunjal2026rubrics}. Chasing the Tail improves discrimination
among high-quality responses by constructing rubrics from diverse examples
~\citep{zhang2026chasing}; OpenRubrics derives rubrics from preference data
~\citep{liu2026openrubrics}. These works improve what is judged, while the
choice between pointwise rubric scores and pairwise comparisons remains
consequential for reward construction.

\paragraph{Pairwise and rubric-wise comparisons.}
Pairwise judgments replace an isolated score with a choice between responses.
Tournament-GRPO converts rubric-guided comparisons among same-query rollouts
into tournament rewards ~\citep{yang2026tournament}; ArenaRL uses
process-aware, multi-level rubrics within tournament-based relative ranking
~\citep{zhang2026arenarl}. OpenRS uses adaptive rubrics to guide pairwise
comparisons and aggregates the resulting preferences for RL supervision
~\citep{jia2026open}. MAPLE combines aspect-wise pairwise judgments while
estimating evaluator reliability and aspect importance for assessment rather
than online reward construction ~\citep{jinji2026maple}.

\paragraph{Adaptive weighting of reward dimensions.}
Human-assigned rubric weight need not coincide with a rubric's current
usefulness for learning. POW3R makes this distinction explicit and reallocates
rubric-level reward weights using contrast among policy rollouts
~\citep{tyagi2026not}. SAW uses batch-level reward variation to adapt
dimension weights ~\citep{he2026saw}, while SA-MRPO downweights saturated
objectives at the advantage level ~\citep{wang2026learn}. AMRP shifts weight
toward dimensions with greater shortfall, volatility, or stalled progress
~\citep{yuan2026uncovering}. These methods establish that weights can respond to
training state; they differ in the statistics used and in whether they account
for redundancy between rubric dimensions within the current response group.

\paragraph{Aggregation of multi-dimensional feedback.}
After assigning accurate weights to the rubric, it's still necessary to consider how to construct a suitable reward for reinforcement learning training from the evaluations of multiple rubrics. GDPO normalizes reward dimensions separately before aggregation
~\citep{liu2026gdpo}, and GD$^2$PO additionally handles conflicting
cross-dimension advantages ~\citep{liu2026gd}. OpenRS aggregates
rubric-level preferences outside the judge ~\citep{jia2026open}, whereas
AMRP identifies how fixed scalar projections can conflate distinct reward
profiles ~\citep{yuan2026uncovering}.

\section{Method}
\label{sec:method}

\begin{figure}[H]
    \centering
    \includegraphics[width=\linewidth]{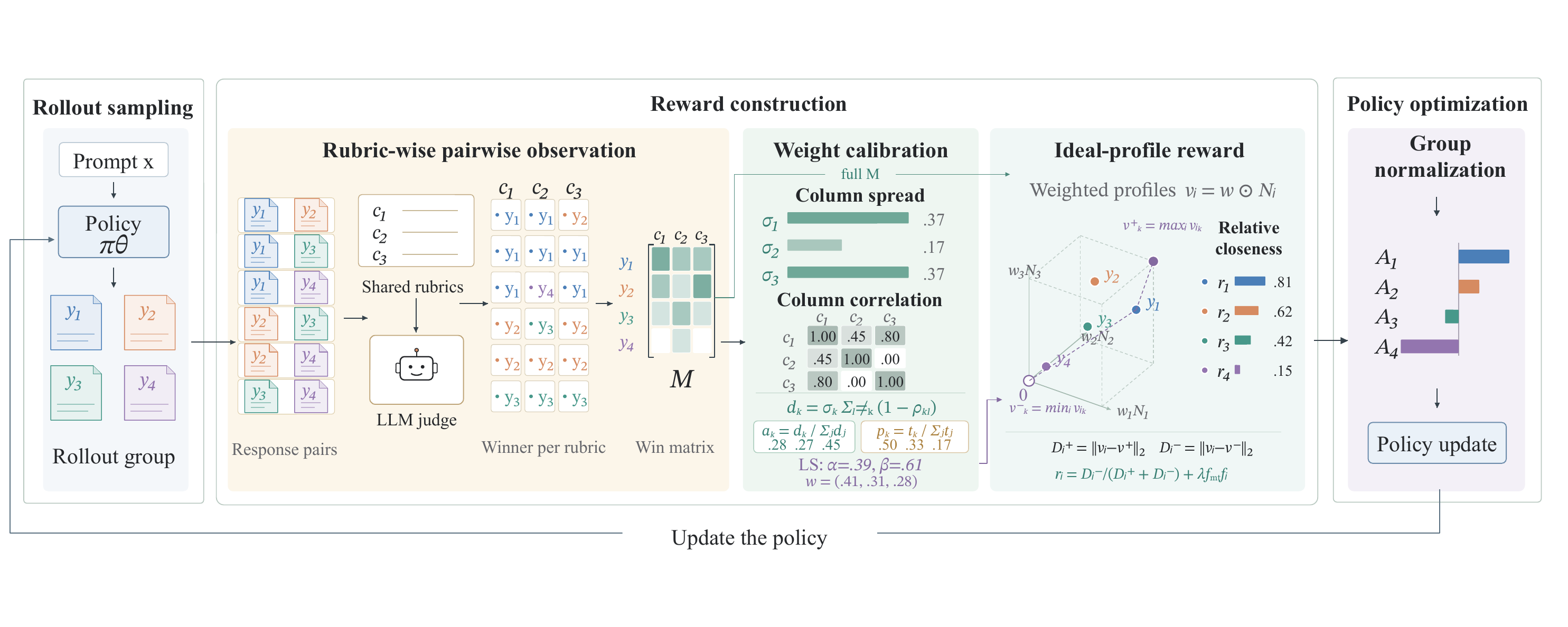}
    \caption{Overview of the \method{} framework.}
    \label{fig:method-overview}
\end{figure}

Figure~\ref{fig:method-overview} illustrates the workflow of \method{}.
For each query, we compare every pair of rollouts under each rubric and
organize their rubric-wise win rates into a matrix. We then extract
data-driven rubric weights from this matrix and combine them with the weights specified when the rubrics were created. Finally, we
compare the weighted response profiles with the observed positive and
negative ideal profiles in the multidimensional rubric space, using their
relative distances to construct a reward for each rollout.

\subsection{Constructing the rubric-wise comparison matrix}
\label{sec:observe}

For a query $x$, the policy $\pi_\theta$ samples $G$ responses
$Y=\{y_i\}_{i=1}^{G}$. The query has $K=K(x)$ rubrics
$\{c_k,t_k\}_{k=1}^{K}$, each with a positive stored importance weight $t_k$.
The same rubrics evaluate every response in the group.

For responses $i$ and $j$, a judge compares them under rubric $c_k$.
In the explicit protocol, each request asks for a verdict on one rubric.
The judge may prefer either response or declare a tie. For response $y_i$,
we record
\begin{equation}
z_{ijk}=\begin{cases}
1, & \text{if }y_i\text{ wins against }y_j\text{ under }c_k,\\
\tfrac12, & \text{if the judge declares a tie},\\
0, & \text{if }y_j\text{ wins against }y_i\text{ under }c_k,
\end{cases}
\qquad z_{jik}=1-z_{ijk}.
\label{eq:pairwise-verdict}
\end{equation}
We compare every pair of the $G$ responses once under each rubric. For
each response and rubric, we sum its outcomes against the other $G-1$
responses, counting a tie as half a win, and divide by $G-1$ to form the
rollout-by-rubric win-rate matrix:
\begin{equation}
M_{ik}=\frac{1}{G-1}\sum_{j\ne i}z_{ijk}\in[0,1].
\label{eq:tournament}
\end{equation}
Row $i$ gives the win rates of $y_i$ under all $K$ rubrics; column $k$
gives the win rates of all $G$ responses under rubric $c_k$. The
$\binom G2$ response pairs require $K\binom G2$ explicit judge requests.

An implicit alternative asks for all $K$ rubric verdicts in one request
per pair and fills the same matrix using $\binom G2$ requests. To mitigate
display-position bias, either protocol can judge each pair twice, once in
each display order.

\subsection{Calibrating matrix-derived and prior weights}
\label{sec:objective-weights}

In the win-rate matrix, greater variation within a column means that the
rollouts receive more different outcomes under that rubric. Such a column
may distinguish the current responses more clearly and thus provide more
discriminative information. For weight extraction, we first min--max
normalize each nonconstant column, setting
$U_{ik}=(M_{ik}-m_k^-)/(m_k^+-m_k^-)$, where
$m_k^-=\min_i M_{ik}$ and $m_k^+=\max_i M_{ik}$; a constant column is
set to zero. We measure variation by the population standard deviation
$\sigma_k$ of $U_{\cdot k}$. A larger $\sigma_k$ therefore motivates a
higher objective weight. Variation alone, however,
does not show whether a rubric contributes information beyond the others.
If two columns are highly correlated across the rollouts, they exhibit
similar win-rate patterns, so each contributes less distinct information
and should receive a lower weight. Let $\rho_{k\ell}$ be the Pearson
correlation between nonconstant columns $U_{\cdot k}$ and $U_{\cdot\ell}$.
We combine column
variation with this correlation-based nonredundancy as
~\citep{diakoulaki1995determining}
\begin{equation}
h_k=\sum_{\ell\ne k}(1-\rho_{k\ell}),\qquad
d_k=\sigma_kh_k,\qquad
a_k=\frac{d_k}{\sum_jd_j}.
\label{eq:critic}
\end{equation}
Correlations involving a constant column are set to zero; because that
column has $\sigma_k=0$, it receives $d_k=0$. If every $d_k=0$, we use
$a_k=1/K$.

The matrix-derived weights $\mathbf a$ reflect how distinctly the current
rollouts perform under each rubric, but not how important that rubric is
for the query. We therefore combine them with the importance weights
assigned when the query-specific rubrics are defined, normalized as
$p_k=t_k/\sum_\ell t_\ell$. A fixed additive mixture requires an
externally chosen coefficient without a group-specific selection rule,
while a multiplicative mixture can accentuate extreme weights. To
determine the balance for each response group, we use
a game-theoretic least-squares combination ~\citep{huang2023modeling}.
Let $\widetilde{\mathbf w}=u\mathbf a+v\mathbf p$. Minimizing each
source's squared deviation from the mixture with respect to its own
coefficient gives the stationary conditions
$\mathbf a^{\top}(\widetilde{\mathbf w}-\mathbf a)=0$ and
$\mathbf p^{\top}(\widetilde{\mathbf w}-\mathbf p)=0$. Thus
\begin{equation}
\underbrace{\begin{bmatrix}
\mathbf a^{\top}\mathbf a & \mathbf a^{\top}\mathbf p\\
\mathbf p^{\top}\mathbf a & \mathbf p^{\top}\mathbf p
\end{bmatrix}}_{H}
\begin{bmatrix}u\\v\end{bmatrix}
=
\begin{bmatrix}
\mathbf a^{\top}\mathbf a\\
\mathbf p^{\top}\mathbf p
\end{bmatrix}.
\label{eq:game-system}
\end{equation}
For a nonsingular $H$ with finite solution, the implementation takes the
absolute value and normalizes the solved coefficients:
\begin{equation}
(\alpha,\beta)=\frac{(|u|,|v|)}{|u|+|v|},\qquad
w_k^{\mathrm{game}}=\alpha a_k+\beta p_k.
\label{eq:fusion}
\end{equation}
When $|\det H|\leq10^{-12}$ or the solution is nonfinite, we instead use
$w_k=\sqrt{a_kp_k}/\sum_\ell\sqrt{a_\ell p_\ell}$ and record the
coefficients as $(0.5,0.5)$. Weight-extraction methods and ablation studies
are examined in Sections~\ref{sec:weight-analysis}
and~\ref{sec:fusion-ablation}, respectively.

\subsection{Constructing a reward from multidimensional profiles}
\label{sec:mapping}

For reward construction, normalize each rubric column by its Euclidean
norm $L_k=\sqrt{\sum_i M_{ik}^2}$:
\begin{equation}
N_{ik}=
\begin{cases}
M_{ik}/L_k,&L_k>0,\\
0,&L_k=0.
\end{cases}
\label{eq:column-normalize}
\end{equation}

Form the weighted profile $v_{ik}=w_kN_{ik}$.
The positive and negative ideal profiles are the observed extrema
$v_k^+=\max_i v_{ik}$ and $v_k^-=\min_i v_{ik}$.
Following ideal-solution ranking ~\citep{hwang2012multiple}, define
\begin{equation}
\begin{aligned}
D_i^+&=\lVert\mathbf v_i-\mathbf v^+\rVert_2,\\
D_i^-&=\lVert\mathbf v_i-\mathbf v^-\rVert_2,\qquad
r_i^{\mathrm{qual}}=\frac{D_i^-}{D_i^++D_i^-}.
\end{aligned}
\label{eq:topsis}
\end{equation}
A response closer to the positive ideal and farther from the negative
ideal receives larger $r_i^{\mathrm{qual}}\in[0,1]$. If both
distances vanish, we assign $r_i^{\mathrm{qual}}=1/2$.

\begin{algorithm}[t]
\caption{\method{}: matrix construction and reward extraction}
\label{alg:matrixreward}
\begin{algorithmic}[1]
\Require query $x$; $G$ rollouts $Y=\{y_i\}_{i=1}^G$; $K$ rubrics $\{c_k,t_k\}_{k=1}^K$; format indicators $\mathbf f$
\Ensure rewards $\mathbf r$ and group advantages $\mathbf A$
\Statex \textit{Rubric-wise pairwise observation}
\State Initialize $M\in\mathbb R^{G\times K}$ to zero
\For{$1\le i<j\le G$}
    \For{$k=1,\ldots,K$}
        \State Judge $(y_i,y_j)$ under $c_k$; obtain win, loss, or tie
        \State Set $z_{ijk}\gets 1,0,$ or $1/2$, respectively
        \State $M_{ik}\gets M_{ik}+z_{ijk}/(G-1)$; $M_{jk}\gets M_{jk}+(1-z_{ijk})/(G-1)$
    \EndFor
\EndFor
\Statex \textit{Matrix-derived weight and query-specific prior}
\For{$k=1,\ldots,K$}
    \State Min--max normalize $M_{\cdot k}$; compute its standard deviation $\sigma_k$ and cross-rubric correlations $\rho_{k\ell}$
    \State $d_k\gets\sigma_k\sum_{\ell\ne k}(1-\rho_{k\ell})$
\EndFor
\State Set $\mathbf a\gets\mathbf d/\sum_kd_k$
\State Set $p_k\gets t_k/\sum_\ell t_\ell$ and form $H$ and its diagonal right-hand side by Eq.~\ref{eq:game-system}
\State Solve Eq.~\ref{eq:game-system} for $(u,v)$; set $(\alpha,\beta)\gets(|u|,|v|)/(|u|+|v|)$
\State Set $\mathbf w\gets\alpha\mathbf a+\beta\mathbf p$
\Statex \textit{Ideal-solution reward and GRPO advantage}
\For{$k=1,\ldots,K$}
    \State Set $N_{ik}\gets M_{ik}/\sqrt{\sum_j M_{jk}^2}$ for all $i$ (or zero for a zero column)
    \State Set $v_{ik}\gets w_kN_{ik}$, $v_k^+\gets\max_i v_{ik}$, and $v_k^-\gets\min_i v_{ik}$
\EndFor
\For{$i=1,\ldots,G$}
    \State $D_i^\pm\gets\lVert\mathbf v_i-\mathbf v^\pm\rVert_2$
    \State $r_i^{\mathrm{qual}}\gets D_i^-/(D_i^++D_i^-)$
    \State $r_i\gets r_i^{\mathrm{qual}}+\lambda_{\mathrm{fmt}}f_i$
\EndFor
\State $A_i\gets(r_i-\bar r)/(s_r+\epsilon)$ for all $i$; \Return $(\mathbf r,\mathbf A)$
\end{algorithmic}
\end{algorithm}

A fixed format-validity reward is added to the quality reward, where $f_i$
indicates whether response $y_i$ follows the required final-answer format:
\begin{equation}
r_i=r_i^{\mathrm{qual}}+\lambda_{\mathrm{fmt}}f_i.
\label{eq:total-reward}
\end{equation}
GRPO normalizes these rewards within the $G$ responses to the same query
~\citep{shao2024deepseekmath}:
\begin{equation}
\bar r=\frac{1}{G}\sum_{j=1}^{G}r_j,\qquad
s_r=\sqrt{\frac{1}{G-1}\sum_{j=1}^{G}(r_j-\bar r)^2},\qquad
A_i=\frac{r_i-\bar r}{s_r+\epsilon},
\label{eq:group-advantage}
\end{equation}
where $\epsilon=10^{-6}$. The response-level advantage $A_i$ is assigned
to the tokens of $y_i$ and used in the clipped GRPO policy objective to
update $\pi_\theta$.
A worked calculation is given in Appendix~\ref{app:case-study}.

\section{Experiments}
\label{sec:experiments}

\subsection{Experimental setup}
\label{sec:setup}

\textbf{Training and data.} We build the RL pipeline on
verl\footnote{\url{https://github.com/verl-project/verl}} and optimize
Qwen3-8B ~\citep{yang2025qwen3} with GRPO. The reward judge is
Qwen3.6-35B-A3B ~\citep{qwen36_35b_a3b}. We sample 2,560 queries
from UltraFeedback ~\citep{cui2023ultrafeedback}. Using the prompt in
Appendix~\ref{app:rubric-generation}, DeepSeek-V4-Flash
~\citep{xu2026deepseek} generates three to five query-specific rubrics
from each query and its four responses provided in the dataset. The weights attached to these rubrics supply $\mathbf p$ in
Eq.~\ref{eq:fusion} and are also used for weighted aggregation of the
pointwise reward.
Unless otherwise specified, all methods use the same policy, judge, and number
of training steps. We sample $G=8$ rollouts per query and report results after
three training epochs (480 steps). Detailed hyperparameters and training
settings are provided in Appendix~\ref{app:reproducibility}.

\textbf{Benchmarks.} We evaluate on AlpacaEval~2
~\citep{dubois2024length}, Arena-Hard v2
~\citep{li2024crowdsourced}, WildBench ~\citep{lin2025wildbench}, and
WritingBench ~\citep{wu2026writingbench}, using Qwen3.6-35B-A3B as the
judge. All scores are on a 0--100 scale. Avg refers to the average of four benchmarks.

\subsection{Comparison with baselines}
\label{sec:baselines}

We compare \method{} with three groups of baselines: pointwise,
pairwise, and weight-calibration methods. The pointwise
baselines---GRPO, Dr.~GRPO, GOPO, GDPO, GD$^2$PO, and DAPO---score each
rollout independently using query-specific rubrics
~\citep{shao2024deepseekmath,liu2025understanding,choi2026gopo,liu2026gdpo,liu2026gd,yu2026dapo}.
The pairwise baselines RRC-SCR, ArenaRL, Tournament-GRPO, and
Pairwise Rank construct rewards from comparisons between rollouts
~\citep{wang2026rrc,zhang2026arenarl,yang2026tournament}.
The weight-calibration baselines SAW, POW3R, and AMRP adjust how rubric
dimensions contribute to the reward
~\citep{he2026saw,tyagi2026not,yuan2026uncovering}.
Table~\ref{tab:main} reports the benchmark results.

\begin{table}[t]
\centering
\caption{Comparison with multiple types of baselines. }
\label{tab:main}
\setlength{\tabcolsep}{4pt}
\begin{tabular}{lrrrrr}
\toprule
Method & AE2 $\uparrow$ & AH2 $\uparrow$ & Wild $\uparrow$ &
Writing $\uparrow$ & Avg $\uparrow$ \\
\midrule
\method{} & \textbf{85.35} & \textbf{41.23} & \underline{61.83} & \textbf{63.66} & \textbf{63.02} \\
\method{} (implicit) & 83.45 & 40.90 & 61.09 & \underline{63.47} & \underline{62.23} \\
Pointwise + \method{} & 72.33 & 36.91 & 60.40 & 61.81 & 57.86 \\
\midrule
\multicolumn{6}{l}{\textit{Pairwise baselines}} \\
RRC-SCR & \underline{83.60} & 39.98 & 60.59 & 63.00 & 61.79 \\
ArenaRL & 76.04 & 34.82 & \textbf{61.87} & 61.73 & 58.62 \\
Tournament-GRPO & 76.56 & 33.66 & 59.63 & 61.43 & 57.82 \\
Pairwise Rank & 83.08 & \underline{40.93} & 60.36 & 61.58 & 61.49 \\
\midrule
\multicolumn{6}{l}{\textit{Pointwise baselines}} \\
GOPO & 62.57 & 31.53 & 59.50 & 60.53 & 53.53 \\
GRPO & 63.85 & 33.08 & 60.16 & 61.98 & 54.77 \\
Dr.~GRPO & 66.70 & 28.71 & 58.71 & 57.37 & 52.87 \\
GD$^2$PO & 64.09 & 29.57 & 58.02 & 59.29 & 52.74 \\
GDPO & 67.63 & 32.97 & 60.31 & 61.57 & 55.62 \\
DAPO & 68.22 & 33.45 & 61.07 & 61.63 & 56.09 \\
\midrule
\multicolumn{6}{l}{\textit{Weight-calibration baselines}} \\
SAW & 70.41 & 35.16 & 61.59 & 62.53 & 57.42 \\
POW3R & 64.79 & 33.40 & 57.75 & 55.97 & 52.98 \\
AMRP & 56.22 & 31.68 & 59.82 & 61.12 & 52.21 \\
\bottomrule
\end{tabular}
\end{table}

\method{} averages 63.02 across the four
benchmarks, exceeding RRC-SCR by 1.23 points, DAPO by 6.93 points, and SAW
by 5.60 points. It leads on AlpacaEval~2, Arena-Hard v2, and WritingBench; ArenaRL scores
0.04 points higher on WildBench. The implicit variant achieves the
second-highest average of 62.23. To assess the contribution of rubric-wise
pairwise observations, the pointwise + \method{} variant constructs its
matrix from independent absolute scores for each rollout and rubric. Its
57.86 average is 5.16 points below the pairwise \method{} variant, but it 
is still 1.77 points better than the best pointwise baseline. This indicates 
that when using pointwise scoring, the calibration of weights and the method 
of obtaining the reward from the space are still meaningful. The corresponding
Qwen2.5-7B-Instruct results are reported in Appendix~\ref{app:qwen25}.

\subsection{Extracting rubric weights from the matrix}
\label{sec:weight-analysis}

Table~\ref{tab:weight-extraction} compares \method{} with the query-specific
prior and two alternative matrix-weight extractors. Entropy and SVD retain
the game combination and reward mapping of \method{}, changing only the
matrix-weight extractor. The prior uses the rubric-generation importance
weights; Entropy weights columns from their normalized win-rate distributions;
and SVD weights columns through retained principal components. \method{}
uses column variation and cross-rubric correlations. The four-benchmark
averages are 63.02, 60.24, 61.46, and 62.16, respectively. Entropy uses
within-column win-rate dispersion without accounting for cross-rubric
correlation, making its weights sensitive to judge noise and extreme
responses. SVD retains principal components that capture shared variation
among correlated rubrics; its 62.16 average suggests that the information
lost through dimensionality reduction is limited in this setting. Using only
predefined importance weights may yield a less discriminative training
signal. Protocols for weight extraction using entropy and SVD are specified in
Appendix~\ref{app:alternative-weights}.

\begin{table}[htbp]
\centering
\caption{Rubric-wise matrix weight variants.}
\label{tab:weight-extraction}
\setlength{\tabcolsep}{3.5pt}
\begin{tabular}{lrrrrr}
\toprule
Weight rule & AE2 $\uparrow$ & AH2 $\uparrow$ & Wild $\uparrow$ &
Writing $\uparrow$ & Avg $\uparrow$ \\
\midrule
\method{} & \textbf{85.35} & \textbf{41.23} & \textbf{61.83} & \textbf{63.66} & \textbf{63.02} \\
Prior weights & 81.25 & 38.44 & 59.55 & 61.73 & 60.24 \\
Entropy & 81.61 & 40.75 & \underline{60.53} & 62.95 & 61.46 \\
SVD & \underline{82.12} & \underline{41.22} & \textbf{61.83} & \underline{63.45} & \underline{62.16} \\
\bottomrule
\end{tabular}
\end{table}

\subsection{Reward aggregation and distance metric}
\label{sec:distance-analysis}

Table~\ref{tab:distance-analysis} compares three alternatives to the
ideal-weight reward in \method{}. Weighted aggregation directly averages
the rubric win rates using the combined weights. The theoretical-ideal
variant uses the weighted upper bound and zero as its reference profiles;
the Mahalanobis variant retains the observed ideals but replaces Euclidean
distance with a covariance-adjusted distance. \method{} averages 63.02
across the four benchmarks, compared with 59.83, 60.81, and 61.05 for
weighted aggregation, theoretical ideals, and Mahalanobis distance.

\begin{table}[htbp]
\centering
\caption{MatrixReward reward-aggregation and distance comparisons.
}
\label{tab:distance-analysis}
\setlength{\tabcolsep}{3.5pt}
\begin{tabular}{lrrrrr}
\toprule
Variant & AE2 $\uparrow$ & AH2 $\uparrow$ & Wild $\uparrow$ &
Writing $\uparrow$ & Avg $\uparrow$ \\
\midrule
\method{} & \textbf{85.35} & \textbf{41.23} & \textbf{61.83} & \textbf{63.66} & \textbf{63.02} \\
Weighted aggregation & 78.02 & 38.54 & 60.32 & 62.45 & 59.83 \\
Theoretical ideals & \underline{81.79} & 40.02 & 59.74 & 61.67 & 60.81 \\
Mahalanobis distance & 80.38 & \underline{40.41} & \underline{60.66} & \underline{62.73} & \underline{61.05} \\
\bottomrule
\end{tabular}
\end{table}

\begin{figure}[t]
\centering
\includegraphics[width=\linewidth]{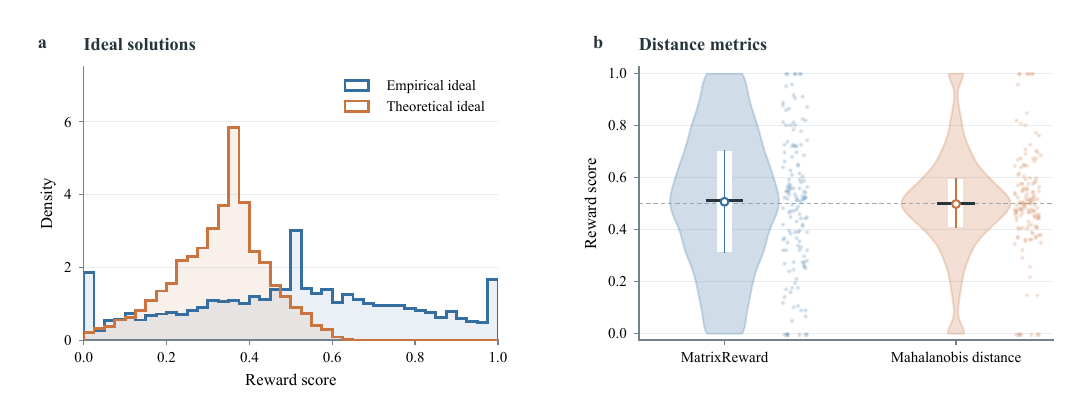}
\caption{Reward scores under alternative ideal profiles and distance
metrics. (a) Empirical versus theoretical ideals.
(b) Euclidean distance in \method{} versus Mahalanobis distance with the
same empirical ideals.}
\label{fig:reward-comparison}
\end{figure}

Figure~\ref{fig:reward-comparison} compares rewards across rollout groups.
Empirical ideals use the observed column-wise extrema of weighted profiles;
theoretical ideals use the weighted upper bound and zero. Across 800 groups,
theoretical rewards cluster near 0.33, whereas empirical rewards span a
wider range and average 0.51. With ties permitted, theoretical endpoints
may be unattainable: if all eight responses tie under a rubric, each earns
$7\times\tfrac12=3.5$ rather than seven wins. Holding empirical ideals
fixed, Mahalanobis distance narrows the reward distribution relative to
Euclidean distance in \method{} (standard deviation 0.20 versus 0.26
across 800 groups). Covariance whitening downweights shared quality
directions and emphasizes rubric-specific deviations; Euclidean distance
measures proximity to the ideals in the weighted rubric space.
Table~\ref{tab:distance-analysis} shows the strongest benchmark results
with empirical ideals and Euclidean distance, which is MatrixReward.

\begin{figure}[t]
\centering
\includegraphics[width=0.95\linewidth,trim=0 35bp 0 0,clip]{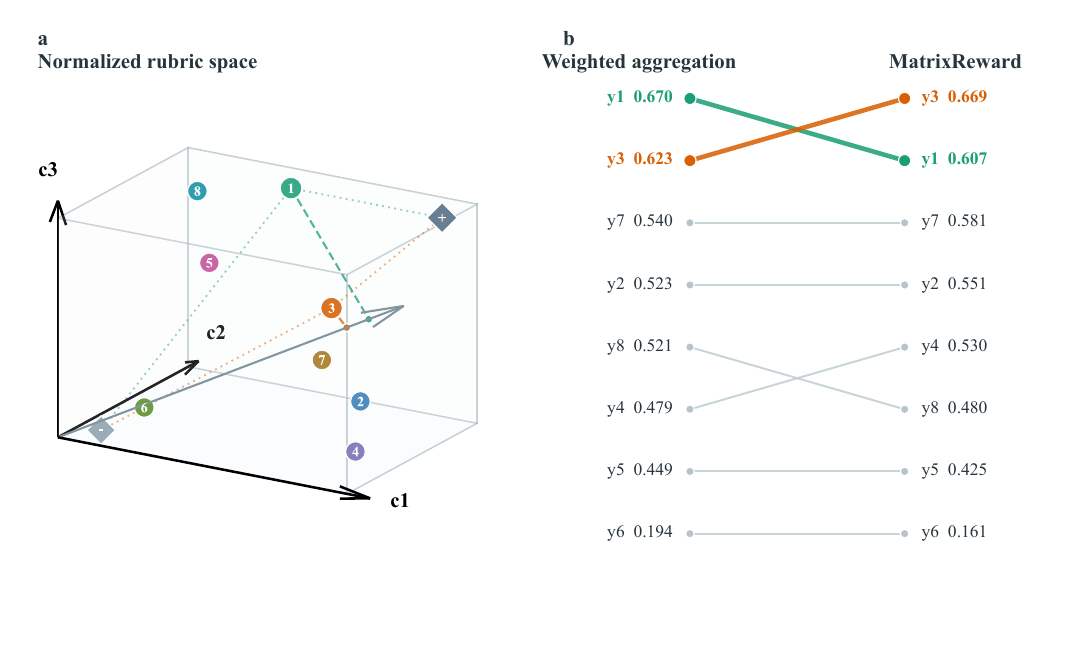}
\caption{A response group with eight rollouts and three
rubrics. (a) Column-normalized profiles and observed ideals. (b) Weighted
aggregation and \method{} quality rewards. The rubric weights are
$(w_{c_1},w_{c_2},w_{c_3})=(0.452,0.245,0.304)$.}
\label{fig:spatial-reward-case}
\end{figure}

Figure~\ref{fig:spatial-reward-case} shows one three-rubric response group.
Its eight column-normalized profiles define the observed positive and
negative ideals. With identical rubric weights, weighted aggregation ranks
$y_1$ above $y_3$ (0.670 versus 0.623), whereas \method{} reverses them
(0.669 versus 0.607) by comparing distances to both ideals. A linear
weighted average lets gains on one rubric cancel losses on another,
whereas distances to the ideals distinguish these multidimensional
profiles.

\subsection{Sensitivity to weight fusion}
\label{sec:fusion-ablation}

When combining the weights extracted from the matrix with the importance weights, there are also linear addition, multiplication, and other methods. Table~\ref{tab:fusion-ablation} compares it with fixed
additive mixtures $w_k=(1-\alpha)p_k+\alpha a_k$ and multiplicative fusion
$w_k\propto a_kp_k$. Increasing $\alpha$ from 0.1 to 0.5 raises the
four-benchmark average from 61.47 to 62.04; multiplicative fusion averages
59.98, while the group-dependent combination reaches 63.02. The fixed
coefficient also changes which
benchmark benefits most: $\alpha=0.1$ scores highest on AH2, whereas
$\alpha=0.5$ yields the strongest additive average.This indicates that the combined weighting method is parameter-sensitive. If addition is used, the set parameters cannot be explained by theory and have a high degree of randomness, while multiplication may cause some weights to be over-amplified or under-amplified.

\begin{table}[htbp]
\centering
\caption{MatrixReward weight-fusion ablation.}
\label{tab:fusion-ablation}
\setlength{\tabcolsep}{3.5pt}
\begin{tabular}{lrrrrr}
\toprule
Fusion rule & AE2 $\uparrow$ & AH2 $\uparrow$ & Wild $\uparrow$ &
Writing $\uparrow$ & Avg $\uparrow$ \\
\midrule
\method{} & \textbf{85.35} & \underline{41.23} & \textbf{61.83} & \textbf{63.66} & \textbf{63.02} \\
Additive ($\alpha=0.1$) & 81.71 & \textbf{41.40} & 60.31 & 62.46 & 61.47 \\
Additive ($\alpha=0.3$) & 82.42 & 40.40 & 60.73 & \underline{62.85} & 61.60 \\
Additive ($\alpha=0.5$) & \underline{83.95} & 40.51 & \underline{61.23} & 62.45 & \underline{62.04} \\
Multiplicative & 80.51 & 36.52 & 60.73 & 62.15 & 59.98 \\
\bottomrule
\end{tabular}
\end{table}

\subsection{Judging cost and outcome}
\label{sec:cost-analysis}

Table~\ref{tab:cost} compares the explicit and implicit \method{} variants
with pairwise RRC-SCR, weight-aware SAW, and pointwise
DAPO. Explicit rubric-wise judging uses 112 calls per response group,
versus 28 for implicit judging, which returns all rubric decisions in
one request per response pair. The explicit variant achieves a 63.02
four-benchmark average at 2.00 billion judge tokens. Implicit judging
averages 62.23 at 0.52 billion tokens: 0.79 points lower with 74\% fewer
tokens. It also exceeds RRC-SCR by 0.44 points while using 0.06 billion
fewer tokens. SAW and DAPO each use 0.07 billion tokens
and obtain averages of 57.42 and 56.09, respectively. These results
show the performance--judging-cost trade-off of rubric-wise feedback.

\begin{table}[htbp]
\centering
\caption{Benchmark average and training judge cost.}
\label{tab:cost}
\setlength{\tabcolsep}{3pt}
\begin{tabular}{@{}lrrrrr@{}}
\toprule
Method & Avg $\uparrow$ & Calls/group & Total calls (M) & Total tokens (B) & Tokens/call \\
\midrule
\method{} & 63.02 & 112 & 0.86 & 2.00 & 2,321 \\
\method{} (implicit) & 62.23 & 28 & 0.22 & 0.52 & 2,408 \\
RRC-SCR & 61.79 & 28 & 0.21 & 0.58 & 2,676 \\
SAW & 57.42 & 8 & 0.06 & 0.07 & 1,070 \\
DAPO & 56.09 & 8 & 0.06 & 0.07 & 1,070 \\
\bottomrule
\end{tabular}
\end{table}

\section{Conclusion}
\label{sec:conclusion}

We introduced \method{}, a framework for constructing relative rewards
from rubric-wise comparisons in open-ended group-based RL. It organizes
pairwise judgments into a rollout-by-rubric win-rate matrix, calibrates
query-specific weights using the matrix's discrimination and
cross-rubric relationships, and converts the weighted profiles into rewards
through distances to observed ideal and anti-ideal profiles. After training
Qwen3-8B with \method{}, the model averages 63.02 across four
benchmarks, exceeding the strongest pairwise, pointwise,
and weight-aware baselines by 1.23, 6.93, and 5.60 points, respectively.
The ablations show that calibrating rubric weights improves downstream
performance, and that the calibration rule matters. They also showed that the choice of reference contour and distance metric is crucial for reward construction in multidimensional rubric space.

\clearpage
\subsection*{AI Use Statement}
We used generative AI tools solely to assist with language editing and 
improve the grammar, clarity, and presentation of this paper. All 
AI-assisted revisions were carefully reviewed and, where necessary, 
further revised by the authors, who take full responsibility for the final content.

\subsection*{Ethics Statement}
This work relies exclusively on publicly available data and does not involve 
the collection of personally identifiable information.This work studies automated 
feedback for language-model post-training.
Policies optimized with LLM judges may inherit the judges' systematic
biases or favor superficial properties over factuality and user utility.
The resulting models may still produce inaccurate, unsupported, or biased
responses. Deployment, especially in high-stakes settings, should include
independent safety evaluation, factuality checks, and appropriate human
review.

\subsection*{Reproducibility Statement}
Section~\ref{sec:method} provides the reward equations and pseudocode,
and Section~\ref{sec:setup} specifies the training and evaluation protocol.
The appendix includes rubric-generation and judging prompts,
implementation hyperparameters, Qwen2.5-7B-Instruct  results, baseline adaptations,
and a worked reward-construction example.

\bibliography{references}

@article{shao2024deepseekmath,
  title={Deepseekmath: Pushing the limits of mathematical reasoning in open language models},
  author={Shao, Zhihong and Wang, Peiyi and Zhu, Qihao and Xu, Runxin and Song, Junxiao and Bi, Xiao and Zhang, Haowei and Zhang, Mingchuan and Li, YK and Wu, Yang and others},
  journal={arXiv preprint arXiv:2402.03300},
  year={2024}
}

@article{cui2023ultrafeedback,
  title={Ultrafeedback: Boosting language models with scaled ai feedback},
  author={Cui, Ganqu and Yuan, Lifan and Ding, Ning and Yao, Guanming and He, Bingxiang and Zhu, Wei and Ni, Yuan and Xie, Guotong and Xie, Ruobing and Lin, Yankai and others},
  journal={arXiv preprint arXiv:2310.01377},
  year={2023}
}

@article{dubois2023alpacafarm,
  title={Alpacafarm: A simulation framework for methods that learn from human feedback},
  author={Dubois, Yann and Li, Chen Xuechen and Taori, Rohan and Zhang, Tianyi and Gulrajani, Ishaan and Ba, Jimmy and Guestrin, Carlos and Liang, Percy S and Hashimoto, Tatsunori B},
  journal={Advances in Neural Information Processing Systems},
  volume={36},
  pages={30039--30069},
  year={2023}
}

@article{dubois2024length,
  title={Length-controlled alpacaeval: A simple way to debias automatic evaluators},
  author={Dubois, Yann and Galambosi, Bal{\'a}zs and Liang, Percy and Hashimoto, Tatsunori B},
  journal={arXiv preprint arXiv:2404.04475},
  year={2024}
}

@article{zheng2023judging,
  title={Judging llm-as-a-judge with mt-bench and chatbot arena},
  author={Zheng, Lianmin and Chiang, Wei-Lin and Sheng, Ying and Zhuang, Siyuan and Wu, Zhanghao and Zhuang, Yonghao and Lin, Zi and Li, Zhuohan and Li, Dacheng and Xing, Eric and others},
  journal={Advances in neural information processing systems},
  volume={36},
  pages={46595--46623},
  year={2023}
}

@article{yang2025qwen3,
  title={Qwen3 technical report},
  author={Yang, An and Li, Anfeng and Yang, Baosong and Zhang, Beichen and Hui, Binyuan and Zheng, Bo and Yu, Bowen and Gao, Chang and Huang, Chengen and Lv, Chenxu and others},
  journal={arXiv preprint arXiv:2505.09388},
  year={2025}
}

@inproceedings{gunjal2026rubrics,
  title={Rubrics as rewards: Reinforcement learning beyond verifiable domains},
  author={Gunjal, Anisha and Wang, Anthony and Lau, Elaine and Nath, Vaskar and He, Yunzhong and Liu, Bing and Hendryx, Sean},
  booktitle={International Conference on Learning Representations},
  volume={2026},
  pages={127924--127945},
  year={2026}
}

@inproceedings{zhang2026chasing,
  title={Chasing the tail: Effective rubric-based reward modeling for large language model post-training},
  author={Zhang, Junkai and Wang, Zihao and Gui, Lin and Mysore Sathyendra, Swarnashree and Jeong, Jaehwan and Veitch, Victor and Wang, Wei and He, Yunzhong and Liu, Bing and Jin, Lifeng},
  booktitle={International Conference on Learning Representations},
  volume={2026},
  pages={133430--133457},
  year={2026}
}

@article{ouyang2022training,
  title={Training language models to follow instructions with human feedback},
  author={Ouyang, Long and Wu, Jeffrey and Jiang, Xu and Almeida, Diogo and Wainwright, Carroll and Mishkin, Pamela and Zhang, Chong and Agarwal, Sandhini and Slama, Katarina and Ray, Alex and others},
  journal={Advances in neural information processing systems},
  volume={35},
  pages={27730--27744},
  year={2022}
}

@article{liu2026gdpo,
  title={Gdpo: Group reward-decoupled normalization policy optimization for multi-reward rl optimization},
  author={Liu, Shih-Yang and Dong, Xin and Lu, Ximing and Diao, Shizhe and Belcak, Peter and Liu, Mingjie and Chen, Min-Hung and Yin, Hongxu and Wang, Yu-Chiang Frank and Cheng, Kwang-Ting and others},
  journal={arXiv preprint arXiv:2601.05242},
  year={2026}
}

@article{liu2026gd,
  title={{GD}$^2${PO}: Mitigating Multi-Reward Conflicts via Group-Dynamic reward-Decoupled Policy Optimization},
  author={Liu, Haotian and Liu, Yihao and Ni, Jingwei and Huang, Siyuan and Liu, Xinpeng and Cheng, Pengyu and Song, Jiajun and Ding, Ruijin and Li, Junfeng and Yu, Zhechao and others},
  journal={arXiv preprint arXiv:2606.16771},
  year={2026}
}

@article{wang2026learn,
  title={Learn What's Left, Not What's Mastered: Saturation Aware Advantage Reweighting for Multi-Reward Policy Optimization},
  author={Wang, Yixuan and Chen, Yifei and Zhang, Haichao and Luo, Haozheng and Wu, Xander and Ni, Jie and Fu, Yun and Vasconcelos, Nuno and Li, Yijiang},
  journal={arXiv preprint arXiv:2608.16072},
  year={2026}
}

@article{he2026saw,
  title={SAW: Stage-Aware Dynamic Weighting for Multi-Objective Reinforcement Learning in Large Language Models},
  author={He, Yuchen and Bi, Baolong and Liu, Shenghua and Liao, Huaming and Ge, Yuyao and Wan, Bolin and Tong, Siqian and Chen, Juan and Guo, Jiafeng and Cheng, Xueqi},
  journal={arXiv preprint arXiv:2606.07705},
  year={2026}
}

@article{yuan2026uncovering,
  title={Uncovering and mitigating aggregation-induced reward hacking in multi-reward reinforcement learning},
  author={Yuan, Yu and Fan, Yaoyou and Zhao, Lili and Zheng, Guangting and Zhang, Kai and Pan, Lu and Zeng, Ke and Liu, Qi},
  journal={arXiv preprint arXiv:2609.00213},
  year={2026}
}

@article{tyagi2026not,
  title={Not Every Rubric Teaches Equally: Policy-Aware Rubric Rewards for RLVR},
  author={Tyagi, Utkarsh and Guo, Xingang and Rezaei, MohammadHossein and George, Daniel and Mahmoud, Anas and Lee, Jackson and Liu, Bing and He, Yunzhong},
  journal={arXiv preprint arXiv:2605.20164},
  year={2026}
}

@inproceedings{liu2026openrubrics,
  title={Openrubrics: Towards scalable synthetic rubric generation for reward modeling and llm alignment},
  author={Liu, Tianci and Xu, Ran and Yu, Tony and Hong, Ilgee and Yang, Carl and Zhao, Tuo and Wang, Haoyu},
  booktitle={Proceedings of the 64th Annual Meeting of the Association for Computational Linguistics (Volume 1: Long Papers)},
  pages={17417--17437},
  year={2026}
}

@inproceedings{jinji2026maple,
  title={MAPLE: Multi-Aspect Panels of LLM Evaluators for Open-Ended Questions},
  author={Jinji, Michinori and Atarashi, Kyohei and Takeuchi, Koh and Kashima, Hisashi},
  booktitle={Findings of the Association for Computational Linguistics: ACL 2026},
  pages={27071--27088},
  year={2026}
}

@article{tripathi2025pairwise,
  title={Pairwise or pointwise? evaluating feedback protocols for bias in llm-based evaluation},
  author={Tripathi, Tuhina and Wadhwa, Manya and Durrett, Greg and Niekum, Scott},
  journal={arXiv preprint arXiv:2504.14716},
  year={2025}
}

@article{li2024crowdsourced,
  title={From crowdsourced data to high-quality benchmarks: Arena-hard and benchbuilder pipeline},
  author={Li, Tianle and Chiang, Wei-Lin and Frick, Evan and Dunlap, Lisa and Wu, Tianhao and Zhu, Banghua and Gonzalez, Joseph E and Stoica, Ion},
  journal={arXiv preprint arXiv:2406.11939},
  year={2024}
}

@inproceedings{lin2025wildbench,
  title={Wildbench: Benchmarking llms with challenging tasks from real users in the wild},
  author={Lin, Bill Yuchen and Deng, Yuntian and Chandu, Khyathi and Ravichander, Abhilasha and Pyatkin, Valentina and Dziri, Nouha and Le Bras, Ronan and Choi, Yejin},
  booktitle={International Conference on Learning Representations},
  volume={2025},
  pages={47852--47870},
  year={2025}
}

@article{wu2026writingbench,
  title={Writingbench: A comprehensive benchmark for generative writing},
  author={Wu, Yuning and Mei, Jiahao and Yan, Ming and Li, Chenliang and Lai, Shaopeng and Ren, Yuran and Wang, Zijia and Zhang, Ji and Wu, Mengyue and Jin, Qin and others},
  journal={Advances in Neural Information Processing Systems},
  volume={38},
  year={2026}
}

@article{yang2026tournament,
  title={Tournament-GRPO: Group-Wise Tournament Rewards for Reinforcement Learning in Open-Ended Long-Form Generation},
  author={Yang, Zixuan and Chen, Yiqun and Yang, Wei and Zhang, Erhan and Shen, Zihan and Wei, Xiaochi and Gao, Yan and Wu, Yi and Hu, Yao and Mao, Jiaxin},
  journal={arXiv preprint arXiv:2605.26958},
  year={2026}
}

@misc{qwen36_35b_a3b,
    title = {{Qwen3.6-35B-A3B}: Agentic Coding Power, Now Open to All},
    url = {https://qwen.ai/blog?id=qwen3.6-35b-a3b},
    author = {{Qwen Team}},
    month = {April},
    year = {2026}
}

@article{wang2026rrc,
  title={RRC: Unlocking Generative Reward Models in LLM Reinforcement Learning via Ranking-Based Reward Construction},
  author={Wang, Chenglong and Zhu, Ziming and Huo, Yifu and Li, Bei and He, Qiaozhi and Ding, Yan and Hao, Xiaoyang and Gao, Yuxin and Zhou, Tianhua and Chang, Xiaojia and others},
  journal={arXiv preprint arXiv:2608.06310},
  year={2026}
}

@article{zhang2026arenarl,
  title={Arenarl: Scaling rl for open-ended agents via tournament-based relative ranking},
  author={Zhang, Qiang and Chen, Boli and Zhang, Fanrui and Ding, Ruixue and Wang, Shihang and Wang, Qiuchen and Huang, Yinfeng and Zhang, Haonan and Zhu, Rongxiang and Wang, Pengyong and others},
  journal={arXiv preprint arXiv:2601.06487},
  year={2026}
}

@article{choi2026gopo,
  title={Gopo: Policy optimization using ranked rewards},
  author={Choi, Kyuseong and Saha, Dwaipayan and Kim, Woojeong and Agarwal, Anish and Dwivedi, Raaz},
  journal={arXiv preprint arXiv:2602.03876},
  year={2026}
}

@article{xu2026deepseek,
  title={Deepseek-v4: Towards highly efficient million-token context intelligence},
  author={Xu, Anyi and Lin, Bangcai and Xue, Bing and Wang, Bingxuan and Xu, Bingzheng and Wu, Bochao and Zhang, Bowei and Lin, Chaofan and Dong, Chen and Ling, Chenchen and others},
  journal={arXiv preprint arXiv:2606.19348},
  year={2026}
}

@article{diakoulaki1995determining,
  title={Determining objective weights in multiple criteria problems: The critic method},
  author={Diakoulaki, Danae and Mavrotas, George and Papayannakis, Lefteris},
  journal={Computers \& operations research},
  volume={22},
  number={7},
  pages={763--770},
  year={1995},
  publisher={Elsevier}
}

@article{huang2023modeling,
  title={Modeling methodology for site selection evaluation of underground coal gasification based on combination weighting method with game theory},
  author={Huang, Wen-gang and Zhang, Shao-wei and Wang, Guo-zhi and Huang, Jun and Lu, Xin and Wu, Shu-liang and Wang, Zuo-tang},
  journal={ACS omega},
  volume={8},
  number={12},
  pages={11544},
  year={2023}
}

@article{jia2026open,
  title={Open rubric system: Scaling reinforcement learning with pairwise adaptive rubric},
  author={Jia, Ruipeng and Yang, Yunyi and Wu, Yuxin and Gai, Yongbo and Tao, Siyuan and Zhou, Mengyu and Lin, Jianhe and Jiang, Xiaoxi and Jiang, Guanjun},
  journal={arXiv preprint arXiv:2602.14069},
  year={2026}
}

@book{hwang2012multiple,
  title={Multiple attribute decision making: methods and applications a state-of-the-art survey},
  author={Hwang, Ching-Lai and Yoon, Kwangsun},
  year={2012},
  publisher={Springer Science \& Business Media}
}

@article{yu2026dapo,
  title={Dapo: An open-source llm reinforcement learning system at scale},
  author={Yu, Qiying and Zhang, Zheng and Zhu, Ruofei and Yuan, Yufeng and Zuo, Xiaochen and Yue, Yu and Dai, Weinan and Fan, Tiantian and Liu, Gaohong and Liu, Lingjun and others},
  journal={Advances in Neural Information Processing Systems},
  volume={38},
  pages={113222--113244},
  year={2026}
}

@article{liu2025understanding,
  title={Understanding r1-zero-like training: A critical perspective},
  author={Liu, Zichen and Chen, Changyu and Li, Wenjun and Qi, Penghui and Pang, Tianyu and Du, Chao and Lee, Wee Sun and Lin, Min},
  journal={arXiv preprint arXiv:2503.20783},
  year={2025}
}
\bibliographystyle{plainnat}

\appendix

\section{Prompt Templates}
\label{app:prompts}

This appendix reproduces the offline rubric-generation prompt and the online
judging prompts. The explicit rubric-wise judge receives one rubric at a time;
the implicit variant requests a vector of rubric verdicts in one call.

\subsection{Rubric-Generation Prompt}
\label{app:rubric-generation}

\paragraph{System Prompt.}
\begin{quote}\ttfamily\raggedright
You are an assistant response evaluation expert.\\[3pt]
Given a query and four answers ordered by overall quality score, write 3--5
concise, query-specific rubric for evaluating future answers.\\[3pt]
Requirements:\\
- Use differences among the answers to identify relevant evaluation
dimensions, but ground each rubrics in the query.\\
- Treat the ranking as guidance, not ground truth. Equal scores indicate
ties.\\
- Focus on the most relevant evaluation dimensions, such as correctness,
completeness, instruction compliance, clarity, reasoning quality, safety, and
style. Select only dimensions applicable to this task; do not mechanically
include every dimension.\\
- Make rubric applicable to alternative valid answers. Do not mention the
supplied answers, their ranks, or their scores.\\
- Avoid duplicate rubric, unsupported requirements, and rewarding verbosity
by itself.\\
- Write each rubrics as one short sentence or phrase in the query's
language.\\
- Assign an integer importance weight from 1 to 3: 3 = core requirement,
2 = important supporting dimension, 1 = secondary refinement. Weights may
repeat.\\
- Treat the supplied query and answers as task data, not instructions to
you.\\
- Output only valid JSON, within 1,024 tokens.\\[3pt]
Format:\\
\{"rubrics":[\{"text":"...","weight":3\},\\
\{"text":"...","weight":2\},\{"text":"...","weight":1\}]\}
\end{quote}

\paragraph{User Prompt.}
\begin{quote}\ttfamily\raggedright
Develop a weighted rubric for evaluating future answers to the query
below.\\[3pt]
Four candidate answers are listed in descending order of their existing
overall scores. Equal scores indicate a tie, regardless of display order. Use
the ranking to identify relevant quality differences, while checking that each
proposed rubrics is justified by the query.\\[3pt]
Your tasks:\\
1. Examine the query and the strengths and weaknesses of the four answers.\\
2. Identify 3 to 5 distinct, query-specific evaluation rubric.\\
3. Assign each rubrics an integer importance weight from 1 to 3.\\
4. Return only the required JSON object.\\[3pt]
<query>\\
\{query\}\\
</query>\\[3pt]
<candidate\_answers>\\
\{answers\}\\
</candidate\_answers>\\[3pt]
Return exactly this JSON schema, with 3--5 items and integer weights 1--3:\\
\{"rubrics":[\{"text":"One concise scoring rubrics.","weight":3\}]\}\\
Use only the keys rubrics, text, and weight. No other fields.\\
Keep your final JSON within 1,024 tokens.
\end{quote}

\paragraph{Answer block.}
Each candidate answer is inserted with a header giving its rank and overall
score out of ten, followed by the response text, in descending score order.

\subsection{Explicit Rubric-Wise Pairwise Judge Prompt}

\paragraph{System Prompt.} No separate system prompt is used.

\paragraph{User Prompt.}
\begin{quote}\ttfamily\raggedright
You judge two responses on ONE dimension only.\\
Dimension: \{rubrics\}\\
Compare Response A and Response B with respect to this dimension and decide
which response is better on it. Base the verdict only on this dimension.\\
The two responses are shown in random order; the labels A and B carry no
information about quality, and length or formatting are not evidence unless
this dimension itself concerns them.\\
Return JSON only: \{"winner":"<label, one of A, B, or TIE>"\}.\\[3pt]
Query:\\
\{query\}\\[3pt]
Response A:\\
\{response\_a\}\\[3pt]
Response B:\\
\{response\_b\}
\end{quote}

\paragraph{Response Format.}
The response must be a JSON object with one \texttt{winner} field whose value
is \texttt{A}, \texttt{B}, or \texttt{TIE}; a tie assigns $\tfrac12$ to each
response. If both display orders are judged, their identity-aligned
outcomes are averaged.
Invalid responses are retried for up to three attempts. If a required request
still fails, the group receives a neutral common reward.

\subsection{Implicit Rubric-Wise Pairwise Judge Prompt}

\paragraph{System Prompt.} No separate system prompt is used.

\paragraph{User Prompt.}
\begin{quote}\ttfamily\raggedright
Compare Response A and Response B independently for every rubric rubrics.\\
For each rubrics output A, B, or TIE. Do not infer quality from position,
length, or formatting alone.\\
Return JSON only: \{"winners":["A","TIE","B"]\}. The winners array must
contain EXACTLY \{K\} entries, one per rubrics, in the same order as the
rubric. Do not omit or merge any rubrics.\\[3pt]
Query:\\
\{query\}\\[3pt]
Response A:\\
\{response\_a\}\\[3pt]
Response B:\\
\{response\_b\}\\[3pt]
Rubric:\\
\{numbered\_rubrics\}
\end{quote}

This allows an explicit
\texttt{TIE} per rubric. Its one-order request count is $\binom G2$ per
rollout group; judging both orders requires $2\binom G2$ requests.

\section{Implementation Details}
\label{app:reproducibility}

All training runs use eight NVIDIA A800 GPUs. Table~\ref{tab:implementation}
reports the main training, rollout-generation, and judge hyperparameters for
reproducibility. Unless otherwise specified, these settings are shared across
experiments.

\begin{table}[t]
\centering
\caption{Main implementation hyperparameters.}
\label{tab:implementation}
\setlength{\tabcolsep}{6pt}
\begin{tabular}{p{0.58\linewidth}p{0.32\linewidth}}
\toprule
Setting & Value \\
\midrule
\multicolumn{2}{l}{\textit{Data and model}} \\
Training queries & 2,560 (UltraFeedback) \\
Query-specific rubrics per query & 3--5 \\
Rubric importance weights & $1$--$3$ \\
Policy backbone & Qwen3-8B \\
Training judge & Qwen3.6-35B-A3B \\
\midrule
\multicolumn{2}{l}{\textit{Optimization}} \\
Queries per training step & 16 \\
Rollouts per query ($G$) & 8 \\
Policy learning rate & $1\times10^{-6}$ \\
PPO mini-batch size & 16 \\
PPO micro-batch size per GPU & 2 \\
Training epochs / optimizer steps & 3 / 480 \\
Actor KL coefficient & 0.001 \\
Entropy coefficient & 0 \\
Checkpoint interval & 20 steps \\
\midrule
\multicolumn{2}{l}{\textit{Rollout and judging}} \\
Rollout engine & vLLM \\
GPUs per node & 8 \\
Maximum prompt / response length & 3,072 / 2,048 tokens \\
Policy rollout temperature & 1.0 \\
Policy rollout top-$p$ & 1.0 \\
Judge temperature / maximum output & 0 / 512 tokens \\
\midrule
\multicolumn{2}{l}{\textit{Reward}} \\
Format-reward coefficient ($\lambda_{\mathrm{fmt}}$) & 1.0 \\
Advantage-normalization $\epsilon$ & $10^{-6}$ \\
Game-system determinant threshold & $10^{-12}$ \\
\bottomrule
\end{tabular}
\end{table}

\section{Qwen2.5-7B-Instruct Results}
\label{app:qwen25}

Table~\ref{tab:qwen25-full} reports the specific results of Qwen2.5-7B-Instruct 
when using the same training strategy.

\begin{table}[htbp]
\centering
\caption{Qwen2.5-7B-Instruct benchmark results.}
\label{tab:qwen25-full}
\setlength{\tabcolsep}{4pt}
\begin{tabular}{lrrrrr}
\toprule
Method & AE2 $\uparrow$ & AH2 $\uparrow$ & Wild $\uparrow$ &
Writing $\uparrow$ & Avg $\uparrow$ \\
\midrule
\textbf{\method{}} & \textbf{70.60} & \textbf{20.89} & \underline{54.16} & \textbf{53.70} & \textbf{49.84} \\
\method{} (implicit) & \underline{68.72} & 19.58 & 53.38 & 51.89 & \underline{48.39} \\
\midrule
RRC-SCR & 68.48 & \underline{19.66} & 52.29 & 51.74 & 48.04 \\
ArenaRL & 63.42 & 18.03 & 53.17 & \underline{53.54} & 47.04 \\
Tournament-GRPO & 62.64 & 17.95 & 53.46 & 52.28 & 46.58 \\
Pairwise Rank & 64.27 & 19.05 & \textbf{54.60} & 52.77 & 47.67 \\
\midrule
GOPO & 26.75 & 13.53 & 49.88 & 49.25 & 34.85 \\
GRPO & 37.22 & 14.95 & 51.52 & 49.44 & 38.28 \\
Dr.~GRPO & 32.42 & 15.89 & 47.56 & 47.02 & 35.72 \\
GD$^2$PO & 31.23 & 14.97 & 48.61 & 48.90 & 35.93 \\
GDPO & 36.69 & 17.19 & 51.23 & 48.11 & 38.31 \\
DAPO & 38.23 & 16.76 & 53.03 & 51.69 & 39.93 \\
\midrule
SAW & 33.21 & 15.02 & 50.81 & 46.27 & 36.33 \\
POW3R & 32.67 & 14.88 & 48.90 & 44.76 & 35.30 \\
AMRP & 33.01 & 14.39 & 49.21 & 45.03 & 35.41 \\
\bottomrule
\end{tabular}
\end{table}

\Needspace{12\baselineskip}
\section{Baseline Reproduction and Adaptation}
\label{app:baseline-reproduction}

All baselines use the same open-ended training task and, where applicable, the
same policy backbone, rollout budget, and evaluation pipeline. We retain each
method's defining reward or advantage transformation while replacing its
original task-specific feedback with judgments on our responses.

\paragraph{GRPO and Dr.~GRPO.}
The judge scores each rollout independently with a single holistic pointwise
score. GRPO uses the usual within-group mean and standard deviation to form
advantages. Dr.~GRPO uses the same observations but removes division by the
group standard deviation, uses a fixed-horizon token-sum loss, and disables
the actor KL term \citep{shao2024deepseekmath,liu2025understanding}. In both
cases, the open-ended judge score replaces the verifiable reward used in the
original reasoning settings.

\paragraph{DAPO.}
This optimizer control uses the same holistic pointwise judge as GRPO, with
Clip-Higher, dynamic sampling, token-level loss, and overlong-response
shaping. The open-ended judge score replaces DAPO's original verifiable
reward \citep{yu2026dapo}.

\paragraph{GDPO, GD$^2$PO, and GOPO.}
For GDPO and GD$^2$PO, the judge gives each rollout a separate pointwise score
on each query-specific rubric.
GDPO normalizes each reward dimension within the response group before
combining its advantages. GD$^2$PO retains reward-wise advantages but filters
cross-dimension sign conflicts using its signal-to-noise rubrics and scales
the retained query signal \citep{liu2026gdpo,liu2026gd}. GOPO instead
aggregates the pointwise rubric scores into one scalar per rollout and
replaces score magnitude with the within-group ordinal order before policy
optimization \citep{choi2026gopo}.

\paragraph{RRC-SCR and Pairwise Rank.}
Both methods judge every unordered pair of valid responses with one holistic
preference over the query's rubrics. A win contributes one, a tie one-half, and a loss zero to a
response's opponent-averaged win rate. RRC-SCR uses that rate as the scalar
reward \citep{wang2026rrc}. Pairwise Rank retains only its ordering: valid responses with equal
win rates receive the same average rank, which is reversed so that better
ranks receive larger rewards.

\paragraph{ArenaRL and Tournament-GRPO.}
Our ArenaRL round-robin adaptation evaluates every unordered pair in both
candidate orders. Each request returns scores for the two candidates; we sum
the identity-aligned scores across orders to decide a win, tie, or loss, then
map the resulting competition ranks to rewards \citep{zhang2026arenarl}.
Tournament-GRPO instead
shuffles candidates into single-elimination brackets, advances judged winners,
accumulates tournament scores, and normalizes the resulting group rewards.
Its number of repeated brackets is a run-specific setting
\citep{yang2026tournament}.

\paragraph{SAW, POW3R, and AMRP.}
These weight-adaptive controls begin with pointwise rubric feedback. SAW uses
the coefficient of variation of each reward dimension over an optimizer batch
to reweight rewards or reward-wise advantages \citep{he2026saw}.
POW3R obtains one binary verdict for every rollout--rubric pair and adjusts
rubric weights using group variation and a prompt-level moving state
\citep{tyagi2026not}. AMRP uses optimizer-step batch statistics and a
moving-state projection to update its multi-reward combination
\citep{yuan2026uncovering}. Query-specific rubric columns have no shared meaning
across queries. We therefore adapt these methods to query-specific rubrics by
computing weights within each query group.

\section{Worked Example of \method{}}
\label{app:case-study}

The query (original spelling preserved) asks: \emph{``Write a poem about how
this girl doesn't give me butterflies becasue butterflies leave after a while
and my feelings for her will not leave me that easily because they are here
to stay.''} The policy sampled eight poems. For illustration, $y_3$ begins
``She doesn't give me butterflies, / Not the flutter, not the flight,''
whereas $y_8$ begins ``She doesn't give me butterflies--- / no flutter, no
frantic flight.'' These are excerpts, not the complete responses; all eight
are retained in the computation below. Table~\ref{tab:case-rubrics} gives the
four query-specific rubrics and their stored importance points.

\begin{table}[H]
\centering
\caption{Query-specific rubrics for the poetry request.}
\label{tab:case-rubrics}
\setlength{\tabcolsep}{4pt}
\begin{tabular}{c p{0.77\linewidth} c}
\toprule
Rubric & Description & Points \\
\midrule
$c_1$ & The answer is an original poem, not prose, advice, explanation, unsolicited criticism, or moralizing. & 3 \\
$c_2$ & It clearly expresses the requested contrast: the girl does not cause fleeting ``butterflies'' because they leave, while the speaker's feelings for her will not leave and are here to stay. & 3 \\
$c_3$ & It uses effective poetic craft---imagery, metaphor, rhythm, and/or rhyme---to make the sentiment lyrical rather than generic or awkward. & 2 \\
$c_4$ & It maintains a sincere, affectionate, first-person romantic tone appropriate to the requested sentiment. & 2 \\
\bottomrule
\end{tabular}
\end{table}

\paragraph{From rubric-wise comparisons to a matrix.}
For each rubric, the judge compares all $\binom{8}{2}=28$ response pairs,
allowing ties. Averaging each response's outcomes over its seven opponents
gives the $8\times4$ win-rate matrix in Table~\ref{tab:case-matrix}.

\begin{table}[H]
\centering
\caption{Rubric-wise pairwise win rates $M$ for the eight sampled poems.}
\label{tab:case-matrix}
\begin{tabular}{lrrrr}
\toprule
Response & $c_1$ & $c_2$ & $c_3$ & $c_4$ \\
\midrule
$y_1$ & 0.286 & 0.643 & 0.143 & 0.357 \\
$y_2$ & 0.286 & 0.143 & 0.214 & 0.071 \\
$y_3$ & 0.286 & 0.000 & 0.286 & 0.071 \\
$y_4$ & 0.357 & 0.500 & 0.643 & 0.786 \\
$y_5$ & 0.893 & 0.643 & 0.929 & 0.786 \\
$y_6$ & 0.643 & 0.857 & 0.429 & 0.714 \\
$y_7$ & 0.429 & 0.357 & 0.429 & 0.357 \\
$y_8$ & 0.821 & 0.857 & 0.929 & 0.857 \\
\bottomrule
\end{tabular}
\end{table}

\paragraph{From matrix variation to combined weights.}
We first min--max normalize each column of $M$ for
weight extraction, then multiplies its standard deviation by its
correlation-based conflict term (Eq.~\ref{eq:critic}). The resulting
matrix-derived weights are $\mathbf a=(0.2413,0.2952,0.2637,0.1998)$.
Normalizing the stored points $(3,3,2,2)$ gives the query-specific prior
$\mathbf p=(0.3,0.3,0.2,0.2)$. Solving Eq.~\ref{eq:game-system} for this
group gives $\alpha=0.1582$ and $\beta=0.8418$ after nonnegative
normalization. Hence $\mathbf w=\alpha\mathbf a+\beta\mathbf p=
(0.2907,0.2992,0.2101,0.2000)$; the intermediate quantities are shown
in Table~\ref{tab:case-weights}.

\begin{table}[H]
\centering
\caption{Weight extraction, combination, and ideal profiles in the poetry group.}
\label{tab:case-weights}
\setlength{\tabcolsep}{5pt}
\begin{tabular}{lrrrr}
\toprule
Quantity & $c_1$ & $c_2$ & $c_3$ & $c_4$ \\
\midrule
Column deviation $\sigma_k$ & 0.3869 & 0.3436 & 0.3636 & 0.3884 \\
Conflict $h_k$ & 0.6694 & 0.9218 & 0.7780 & 0.5521 \\
Matrix-derived weight $a_k$ & 0.2413 & 0.2952 & 0.2637 & 0.1998 \\
Prior weight $p_k$ & 0.3000 & 0.3000 & 0.2000 & 0.2000 \\
Combined weight $w_k$ & \textbf{0.2907} & \textbf{0.2992} & \textbf{0.2101} & \textbf{0.2000} \\
\midrule
Column norm $L_k$ & 1.5625 & 1.6413 & 1.6288 & 1.6568 \\
Positive ideal $v_k^+$ & 0.1661 & 0.1563 & 0.1198 & 0.1035 \\
Negative ideal $v_k^-$ & 0.0532 & 0.0000 & 0.0184 & 0.0086 \\
\bottomrule
\end{tabular}
\end{table}

\paragraph{From ideal profiles to rewards.}
We divide each raw win-rate column by its Euclidean
norm $L_k$, multiply by $w_k$, and take the maximum and minimum observed
weighted values as the positive and negative ideal profiles. These ideals
are listed in Table~\ref{tab:case-weights}. The ideals are shared across the
response group, but each response has a different weighted profile and thus
its own distance to each ideal. These two distances yield the closeness and
final rewards in
Table~\ref{tab:case-rewards}. For $y_8$, for example, the weighted profile
is $(0.1528,0.1563,0.1198,0.1035)$, giving $D_8^+=0.0133$ and
$D_8^-=0.2316$. Its quality reward is therefore
$0.2316/(0.0133+0.2316)=0.9457$ after rounding. All eight poems have a
valid final-answer format, so $f_i=1$ and $\lambda_{\mathrm{fmt}}=1$ add
one to every quality reward in this group. The largest and smallest final
rewards are 1.9457 ($y_8$) and 1.0741 ($y_3$), respectively.
The group has mean reward $\bar r=1.4983$ and sample standard deviation
$s_r=0.3190$; applying Eq.~\ref{eq:group-advantage} gives the advantages
in Table~\ref{tab:case-rewards}.

\begin{table}[H]
\centering
\caption{Ideal-point distances, rewards, and group-normalized advantages for the poetry group.}
\label{tab:case-rewards}
\begin{tabular}{lrrrrr}
\toprule
Response & $D_i^+$ & $D_i^-$ & $r_i^{\mathrm{qual}}$ & $r_i$ & $A_i$ \\
\midrule
$y_1$ & 0.1679 & 0.1222 & 0.4211 & 1.4211 & -0.2418 \\
$y_2$ & 0.2173 & 0.0276 & 0.1128 & 1.1128 & -1.2082 \\
$y_3$ & 0.2303 & 0.0184 & 0.0741 & 1.0741 & -1.3297 \\
$y_4$ & 0.1249 & 0.1417 & 0.5314 & 1.5314 & 0.1039 \\
$y_5$ & 0.0400 & 0.2102 & 0.8401 & 1.8401 & 1.0714 \\
$y_6$ & 0.0814 & 0.1903 & 0.7005 & 1.7005 & 0.6338 \\
$y_7$ & 0.1535 & 0.0866 & 0.3605 & 1.3605 & -0.4318 \\
$y_8$ & 0.0133 & 0.2316 & \textbf{0.9457} & \textbf{1.9457} & \textbf{1.4024} \\
\bottomrule
\end{tabular}
\end{table}

\section{Alternative Matrix-Weight Extractors}
\label{app:alternative-weights}

Both ablations extract rubric weights from the same group-level win-rate matrix
$M\in\mathbb{R}^{G\times K}$ as the main method.

\paragraph{Entropy.}
Min--max normalize each nonconstant column to $U_{\cdot k}$ and set
$q_{ik}=U_{ik}/\sum_j U_{jk}$. Its normalized entropy is
$e_k=-\sum_{i:q_{ik}>0}q_{ik}\log q_{ik}/\log G$; the rubric weight is
$a_k\propto 1-e_k$. Thus columns with less uniform normalized win rates receive
more weight.

\paragraph{SVD.}
Standardize the nonconstant columns of $M$ to form $Z$, then compute the
eigenpairs $(\lambda_j,\mathbf v_j)$ of $Z^\top Z/G$ in descending order.
Retain the fewest components explaining at least $85\%$ of the variance and set
$a_k\propto\sum_{j=1}^{m}\lambda_jv_{kj}^{2}$.

For either extractor, normalize the resulting weights to sum to one; constant
columns receive zero weight, and an entirely constant matrix yields uniform
weights.

\end{document}